\documentclass[10pt,letterpaper]{article}

\usepackage{cogsci}

\cogscifinalcopy 

\usepackage{pslatex}
\usepackage{apacite}
\usepackage{float} 

\usepackage{graphicx}
\usepackage{enumitem}

\title{Do Time Constraints Re-Prioritize Attention to Shapes \\During Visual Photo Inspection?}
 
\author{{\large \bf Yiyuan Yang*\footnote{*Indicates Equal Constribution} (yiyuan.yang@vanderbilt.edu)} \\
  Department of Computer Science, 2301 Vanderbilt Pl \\
  Nashville, TN 37235 USA
  \AND {\large \bf Kenneth Li* (kenneth.li@vanderbilt.edu)} \\
  Department of Computer Science, 2301 Vanderbilt Pl \\
  Nashville, TN 37235 USA
  \AND {\large \bf Fernanda Eliott (eliottfe@grinnell.edu)} \\
  Department of Computer Science, 1116 8th Ave \\
  Grinnell, IA 50112 USA
  \AND {\large \bf Maithilee Kunda (mkunda@vanderbilt.edu)} \\
  Department of Computer Science, 2301 Vanderbilt Pl \\
  Nashville, TN 37235 USA}

\begin{document}

\maketitle


\begin{abstract}
People's visual experiences of the world are easy to carve up and examine along natural language boundaries, e.g., by category labels, attribute labels, etc.  However, it is more difficult to elicit detailed visuospatial information about what a person attends to, e.g., the specific shape of a tree.  Paying attention to the shapes of things not only feeds into well defined tasks like visual category learning, but it is also what enables us to differentiate similarly named objects and to take on creative visual pursuits, like poetically describing the shape of a thing, or finding shapes in the clouds or stars.  We use a new data collection method that elicits people's prioritized attention to shapes during visual photo inspection by asking them to trace important parts of the image under varying time constraints.  Using data collected via crowdsourcing over a set of 187 photographs, we examine changes in patterns of visual attention across individuals, across image types, and across time constraints.

\end{abstract}

\section{Introduction}

\begin{figure}[t]
    \includegraphics[width=0.513\linewidth]{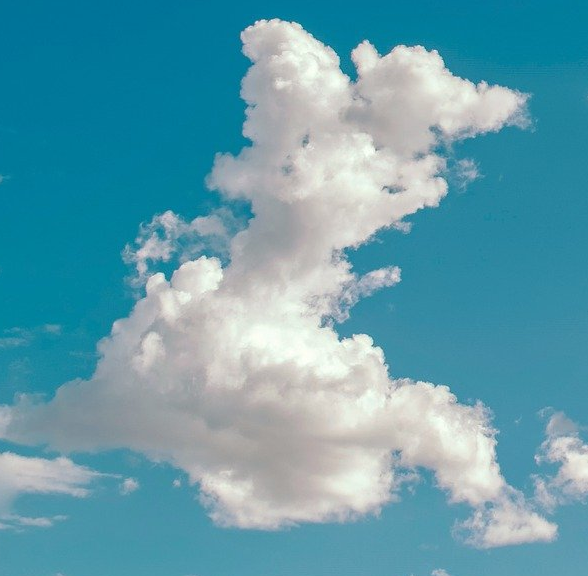} \includegraphics[width=0.47\linewidth]{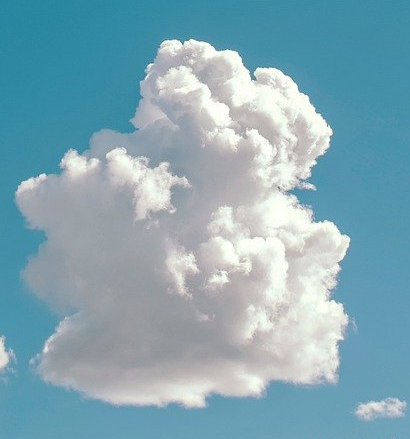}
    \caption{Two clouds.  What do you see?}
    \label{fig:clouds}
\end{figure}

Consider the two clouds shown in Figure \ref{fig:clouds}.  In many ways, these images are the same: they each show one main cloud, with some wispy bits in the corners.  The clouds are pretty much the same color, the background is the same color, and we might describe both clouds as ``fluffy, white, shiny clouds with grey undersides and heads slightly cocked to the right.''

However, the shape of each cloud is clearly different; we would not easily confuse the two visually.  We might describe the left one as more concave and elongated, but no number of words can fully describe the visuospatial differences that we perceive with our eyes.  And, these \textit{highly perceptible} but \textit{hard to verbalize} differences are what fuel our creative games of seeing ``things'' in clouds.  The left could perhaps be Piglet pouring tea, and the right?  A teddy bear on a...moped?


Just as we use language to understand and express ideas verbally, we use shapes (among many other kinds of visual information) to interpret and recreate our visual world.  While early theories of visual perception tried to develop fairly compact shape vocabularies \shortcite{biederman1987recognition}, we expect that our ``shape vocabularies'' are much, much richer, containing commons elements drawn from the complex, messy, and hard-to-verbalize visual world we experience.



In this paper, we examine methods for eliciting and analyzing the shapes that people intuitively attend to in visual photographs.  Specific contributions of this paper include:
\begin{enumerate}[nolistsep,noitemsep]
    \item We describe a new data collection method that asks crowdsourced workers to trace the important shapes within a given image, and we vary the time constraints available on each trial to force our sketchers to prioritize which shapes they choose to draw.
    \item We deploy this method over a dataset of 187 photographs of ``interestingly'' shaped objects from a number of different categories.
    \item We analyze people's distribution of spatial attention over time, as measured by their sketch drawings, and we show that the prioritization of certain shapes becomes increasingly varied under time pressure.
\end{enumerate}

\section{Background}

\textbf{Attention. }
Triesman’s pioneering work established that visual attention can be modeled in two stages, with a period of parallel processing before serialized focused attention gets deployed \shortcite{TREISMAN1986}.  Later researchers have put significant effort into discovering the mechanisms that affect how people process a scene in parallel, and what factors and stimuli influence their formulation of an attention deployment queue after preattentive processing ends. 

Hoffman alluded to the importance of shape properties, which includes relative proportion, protrusion and obvious boundary, for object-part saliency \shortcite{hoffman1997salience}. The concept of perceptual grouping \shortcite{Yantis1992}, i.e, regarding groups of components and shapes as constituting a higher order object, also invited researchers to establish and investigate Gestalt principles and how they affect attention deployment \shortcite{Ojha2013, Marini2016}. Some also looked into how shape plays a role in object-based attention mechanisms \shortcite{wilder2016role}. 

Wolfe summarized the factors by which human attention is affected and while he conceded that 'shape' still lacked a concrete definition, he included it as one of the potentially important visual stimuli in pre-attentive stages \shortcite{Wolfe2017}. More generally, certain shapes are thought to act as hotspots which crave attention in a scene, and some like Hoffman have made their own studies into what can affect how people prioritize the shapes they perceive. 

\textbf{Curve Tracing Attention. }
Directly studying the shapes people perceive has been challenging because shape is a vaguely-defined concept and most researchers either refrain from using complex shapes in their analysis \shortcite{Larson2011} or study shapes without directly identifying them \shortcite{Alexander2014}. 

Line tracing or curve tracing attention has been regarded as one type of attentive processing that takes place during object recognition and serves as a substitute for direct shape-based analysis \shortcite{Jolicoeur1986, Scholte2001}. Work by Jollicoeur and Crundall showed that people sometimes deploy attention along boundary lines in an image or a scene \shortcite{Jolicoeur1991a}, and follow through the line from start to end \shortcite{Crundall2008}.

\textbf{Tracking Attention. }
Eye tracking has been one approach to follow visual attention across time as it can help determine where people lend their focus \shortcite{Liu2011, Tsai2012}. However, eye tracking can also be noisy as it includes periods of times where people are not actually focusing on anything in particular. Other approaches like \shortcite{Freeman2011, Smucker2014} positional recording of a computer mouse have been able to produce higher-quality signals for tracking attention. 

There is also heavy coordination between hand movements and visual attention involved during sketching. Recording the cursor positions of  participants who perform curve tracing over an online interface leverages the tight coupling between visual attention and motor control to track which contours people are attending to. Using sketch in our study also has the benefit of forcing subjects to focus on shapes in a scene or image rather than on colors or textures. 

\textbf{Sketch Datasets. }
Sketch datasets have been created for various computer vision tasks such as sketch recognition,  generation \shortcite{ha2018}, and sketch-based image retrieval \shortcite{Xu2020DeepLF, eitz2012hdhso}. Most datasets like QuickDraw \shortcite{Quickdraw} collect free-hand sketches from people given only verbal category prompts, e.g. ``squirrel.'' 

Since it is difficult to quantify how closely an abstract sketch resembles the target object, there is immense diversity in the sketches produced for open categories like ‘cat’. Attempts to handle the high variation in abstract drawings included manually annotating each stroke with a semantic label \shortcite{li2018universal}. While the data related to the timing and placement of strokes can still be leveraged in a limited capacity to estimate where attention is being deployed, it is still hard to compare sketches because there is immense artistic variation within broad categories. 

Other datasets like QMUL Shoe \shortcite{yu2016sketch} and Sketchy \shortcite{Sangkloy2016} have collected more fine-grained sketches by presenting a source image, like a single image of a specific shoe, and asking users to sketch the target object with a similar pose. However, these schemes show the image only briefly and have users sketch from memory, since the intended purpose is sketch-based retrieval, which treats sketches as queries. In our study, participants directly trace an image which is visible for the entire duration of their input. To avoid intractable variance in sketches, we impose time constraints so that the amount of detail is limited, thus pointing our study to the effect of time pressure on potential re-prioritization of shape attention.

\begin{figure*}[t!]
    \includegraphics[width=\textwidth]{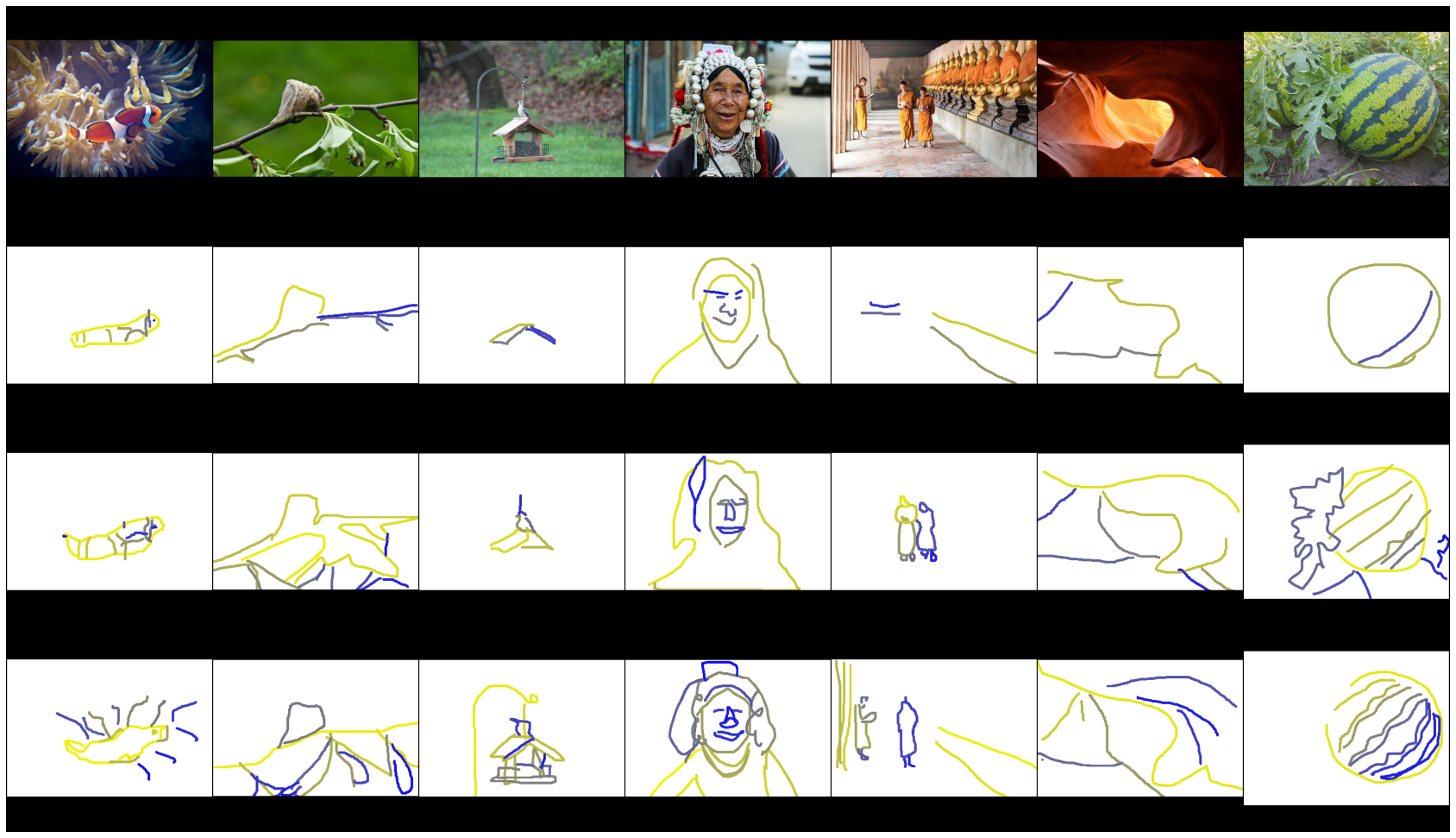}
    \caption{Sample source images and sketches. The ordering of strokes is represented by a shift of color from yellow(earlier) strokes to blue(later) strokes. From top to bottom: a source image, a 10-second sketch, 20-second sketch, and 40-second sketch. }
    \label{fig:Dataset Preview}
\end{figure*}

\section{Methods}

\textbf{Picking Source Images. }
Source images are selected to explore a wide range of shapes that may elicit re-prioritization in attention. Source images are what drawers reference when creating their sketches. Most source images chosen for our dataset are centered on a specific object. 

Datasets like QMUL Shoe/Chair \shortcite{yu2017sketchx}, QMUL Handbag \shortcite{song2017deep}, Swire \shortcite{huang2019swire} Aerial-SI \shortcite{jiang2017sketch} only use source images from a single category. These datasets are useful for fine-grained sketch representations, but the results are not widely applicable to other settings due to being domain-specific. Datasets like Sketchy \shortcite{Sangkloy2016} or TU-Berlin \shortcite{eitz2012hdhso} selected source images from a broader variety of categories and produced findings with wider applicability.

Our source images come from the following super-categories, with the number of images shown in parenthese: \emph{animals} (20), \emph{artifacts} (42), \emph{faces} (20), \emph{foods} (20), \emph{geological formations} (20), \emph{natural objects} (5), \emph{people} (20), \emph{plants} (20), and \emph{abstract art} (20). In addition, selected images not only include those that are familiar to people but also those that are not frequently seen in real life. The source images under the \emph{faces} and \emph{people} categories are selected with ethnic, gender and age diversity in mind. The final source image set includes 187 images under 9 categories.

\subsection{Sketch Requirements}
With the goal of shape identification in mind, our dataset was constructed under the following tenets: 1) Drawers should only draw the main object, and express the contours and shapes as much as possible without any coloring. 2) Drawing is time constrained. 3) Drawers’ shape recognition and prioritization during drawing is recorded and representative.

In most cases such as QuickDraw, QMUL Shoe and Chair, etc., free-hand sketch prompts ask users to produce abstract drawings of objects.  An alternative approach to abstract sketch collection is image tracing, where users are asked to trace the outlines and definitive contours of an object. The image tracing scheme inherently turns user attention towards specific contours of interest while also reducing variation in sketch quality. Additionally, with the precision required to trace curves from photos, visual search is limited in a serial, deliberate capacity and so stroke order can also mirror attentional priority. We leverage image tracing coupled with various time constraints in order to collect a dataset of sketches that can be used to highlight variations in attended shapes.

\subsection{Other Factors}
Studies have shown that various factors like color and spatial location influence people’s attention prioritization \shortcite{Wolfe2014}. Our study is different in that we ask participants to produce contour and shape based sketches. Each participant's goal is to create drawings that others can recognize. Drawers are incentivized to include the most shape-salient, rather than, for instance, color-salient parts of the object, to maximize sketch recognizability. Parts that have the most conspicuous colors or textures may not be the parts that are most useful in a simple sketch. 

\begin{figure*}
    \includegraphics[width=\textwidth]{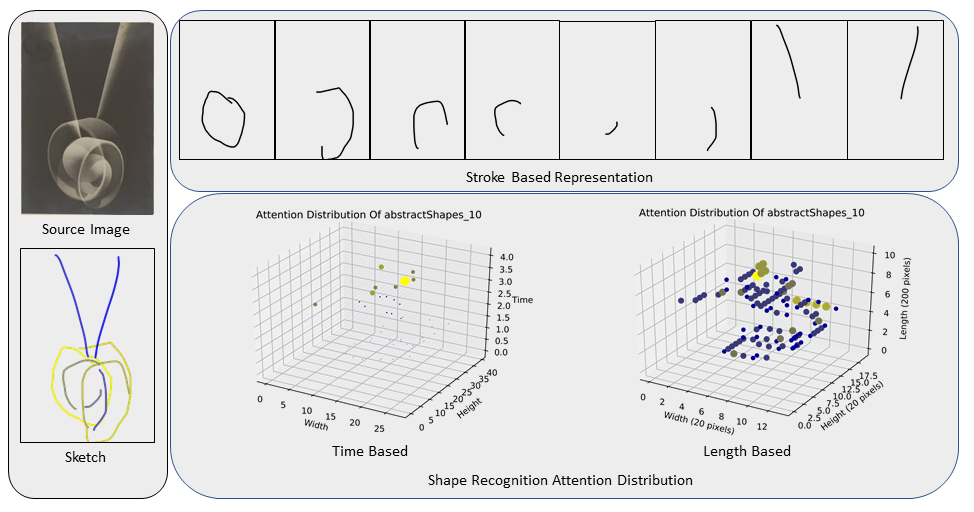}
    \caption{Representing A Traced Sketch}
    \label{fig:Sketch Representation}
\end{figure*}

\subsection{Ensuring Data Quality}
There are several challenges in achieving the first two tenets. First, Amazon Mechanical Turk workers have the incentive to finish as many tasks as possible. Thus, many often include few details and submit the drawing early, which could result in drawing behaviors being misrepresented. Similarly, prompt instructions were often glossed over and several participants likely proceeded without fully understanding the requirements. This initially resulted in a low usability rate across collected data. Therefore, to ensure a higher usability rate and improve workers' understanding of our tasks, we included images of completed drawing samples to assist with producing a completed and high-quality sketch.

\subsection {Time Constraints}
Prior datasets apply time constraints only during source image observation or not at all. A lack of time constraints leaves significant room for variance in the details of sketches. Some people may want to include every detail while others include few, and some people may want to spent more time drawing while others try to finish it quickly. 
Furthermore, the lack of time constraints would also mean the labels (sketches) have relatively weaker supervision since a greater number of factors could have influenced final drawings.

The use of time constraints not only allows our dataset to be used for more rigorous attention prioritization analysis, but also for use cases such as modeling drawing behaviors.

In this experiment, we selected 10 seconds, 20 seconds, and 40 seconds as the three time constraint groups with 20 seconds being the control group, and we only show the source image after the timer starts. With the hypothesis that people re-prioritize shape attention under different time constraints, we observe if a decision to re-prioritize is forced given a sufficiently restrictive time limit. During our initial experiments, we found that a 10 second limit virtually forces drawers to focus on a few parts of the source image since an elaborate sketch is impossible to produce under such a short time frame. Drawings from 20 second group and 40 second group are similar in what they depict, however, the lines drawn in 20 second drawings are usually more hastily done, suggesting users had less time to decide.

\subsection{Data Representation}
Shape 'priority' can either be measured as the length of time used to sketch it, or how much earlier its corresponding strokes precede other strokes. There are several ways to implement these ideas: 1) Treat individual strokes as identifiers of actual shapes in the source images and use the stroke ordering as people's shape centered attention prioritization. 2) Generate a 3D mapping where values x, y, z along each axis means the mouse is at x, y position on the image at time z. 3) Generate a 3D mapping where values x, y, z along each axis means the mouse is at x, y position on the image after drawing for distance z pixels. 

Traditional stroke-based representations may produce the most interpretable results, but people have different preferences on stroke length. While some people may draw the entire outer boundary of the object in a single stroke, others may break it down into multiple strokes. If we strictly compare their first strokes, then second, and then third, then stroke-based comparisons would suggest that they have different preferences even if their strokes align perfectly.

Time-based mouse tracking is ideal for determining where people looked at and for how long. However, our study focuses on attention ordering and not the time spent on them. Two visually indistinguishable images from two different drawers who allocated time differently are considered identical by the attention prioritization standard. 

Therefore, this study uses the third method, which records the mouse's location to effective distance traveled since it strictly compares the order in which people attended and recognized the shapes in a source image. We denote this as a length-based 3D mapping of a sketch. 

We trade off pixel-level granularity of our raw sketch data to simplify the final mappings so that the resulting data representation can be interpreted as a grid of voxels, each of which represents the collective 'ink', or number of pixels, within a \(60_{px} \cdot 60_{px} \cdot 300_{px}\) slice of the length-based 3D mapping. 

\subsection{Collected Dataset}
For each source image out of the 187, and under each of the 10, 20 and 40 second time constraints, we asked 10 drawers to draw the main object on top of the source image, producing 5610 sketches in total. We also collected another 1870 sketches for each 20 second time constraint group in order to perform baseline comparisons. 

\begin{figure*}
    \includegraphics[width=\textwidth]{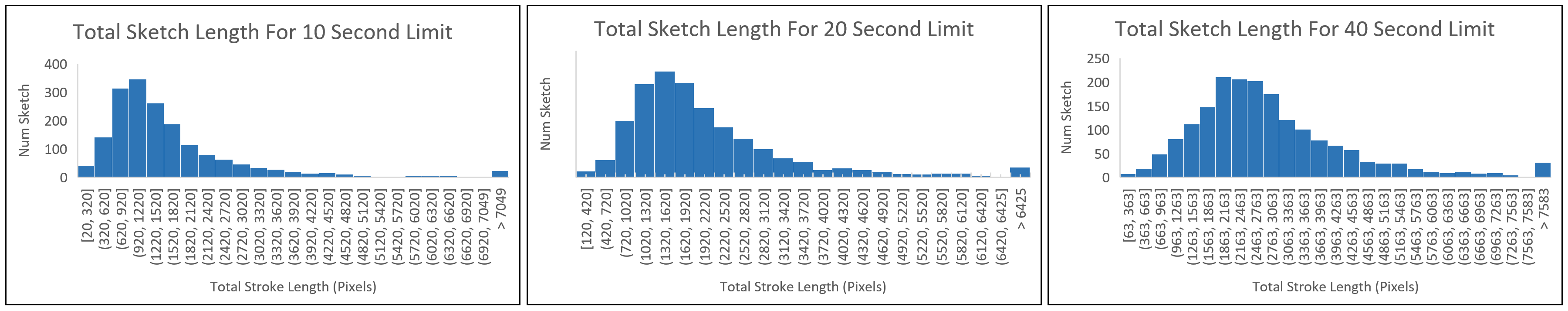}
    \includegraphics[width=\textwidth]{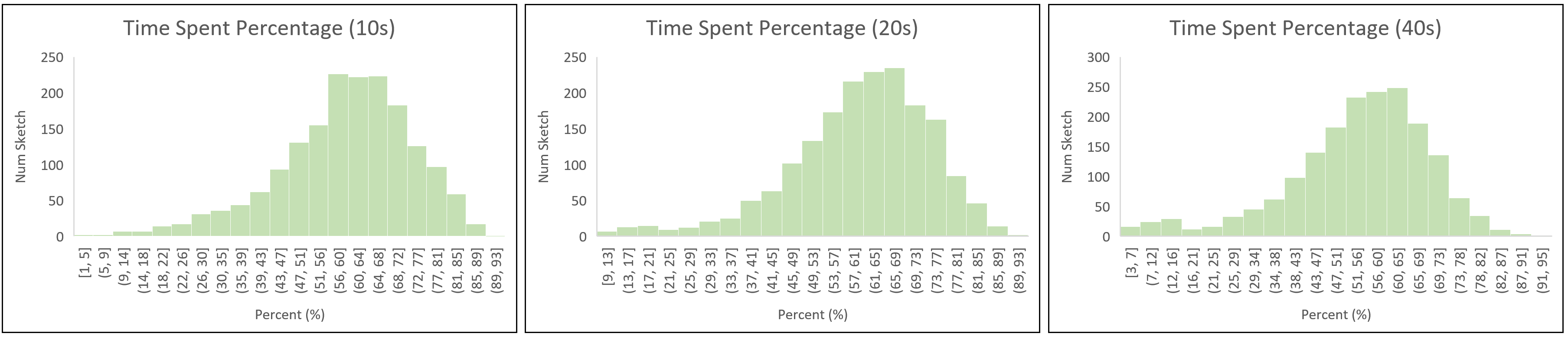}
    \caption{Sketch length (top) and time spent as a percentage of allotted time (bottom) per time constraint group.}
    \label{fig:Sketch Lengths}
\end{figure*}

\subsection{Experiments and Comparison Methods}
To test whether sketches from a time constraint group $A$ prioritized shapes differently compared to those under a time constraint $B$, we first perform an "AA" test. Here, we compare the two sets of sketches within the same 20 second time constraint group. For each source image, we have two sets of 10 sketches for the AA test. We can calculate value \(D_{ij}\) for each sketch pair, which represents the average Euclidean distance between each attention distribution voxel. 
\[D_{ij} = \frac{(X_i - Y_j)^2}{WHL_{ij}} \]
Where \(X_i\) comes from set \(X\) and \(Y_j\) comes from set \(Y\) and the two images have same width and height of \(W\) and \(H\) and have common drawn pixel length of \(L_{ij}\).

This represents the per-voxel variation between two sketches. And the two 10 sketch sets produces 100 possible pairs and thus 100 \(D\) values that allow us to calculate the mean and variance of the \(D\) values for these two sets. Therefore for each source image(\(s\)), we have:
\[D_s = \sum_{i=1}^{10}\sum_{j=1}^{10}\frac{(X_i - Y_j)^2}{WHL_{ij}}\]

Then, in an "AB" test, we would calculate the \(D\) values of the 10 images under 20 seconds against those under 10 seconds or those under 40 seconds. We can then perform a student’s t test with 95\% confidence to reject or fail to reject the null hypothesis that people re-prioritize under time constraints.

Considering that drawers usually produce more elaborated drawings with more time, and a more elaborate drawing's 3D attention distribution mapping is unlikely to be considered similar to those that are simpler, we cut off the extra length in the longer drawings for fair comparison.

\begin{table}[!ht]
\begin{center} 
\caption{Comparing 20s Against 10s} 
\setlength\tabcolsep{2pt}
\begin{tabular}{c c c}
\hline
Category &  Rejections & Rejection Rate \\
\hline
Abstract Shape & 6 & 0.3\\
Animal & 4 & 0.2\\
Artifacts & 12 & 0.29\\
Faces & 8 & 0.4\\
Food & 7 & 0.35\\
Geological Formation & 8 & 0.4\\
Natural Objects & 1 & 0.2\\
People & 4 & 0.2\\
Plant & 8 & 0.4\\
\hline
\end{tabular} 

\caption{Comparing 20s Against 40s} 
\setlength\tabcolsep{2pt}
\begin{tabular}{c c c c} 
\hline
Category &  Rejections & Rejection Rate \\
\hline
Abstract Shape & 2 & 0.1\\
Animal & 3 & 0.15\\
Artifacts & 11 & 0.26\\
Faces & 5 & 0.25\\
Food & 4 & 0.2\\
Geological Formation & 7 & 0.35\\
Natural Objects & 0 & 0\\
People & 2 & 0.1\\
Plant & 5 & 0.25\\
\hline
\end{tabular} 
\end{center} 
\end{table}

\section{Results}
In our experiments, for each source image, we made a separate hypothesis that time constraints would not affect people’s prioritization of attention towards shapes. When comparing against 10 second time constraint, out of 187 source images we rejected hypotheses for 58 and when comparing against 40 second time constraint, we rejected 39. We observe that for every single super category, the shape attention re-prioritization rate is reduced when testing against 40 second time constraint compared to testing against 10 second time constraint. Table 3 shows that there are two times more source images with which we saw shape attention re-prioritization between 10 seconds  and 20 seconds but no attention re-prioritization between 20 seconds and 40 seconds, than the reverse. This shows that when time is more relaxed and the time constraint becomes less relevant as people can finish the drawing under the constraint, there is less re-prioritization of attention towards shapes. In comparison, when time becomes more constrained, and people can no longer finish the drawing under time constraint, there are more instances of attention re-prioritization.  

\section{Conclusion}
To explore the nature of shape saliency in visual attention, we exploit hand-drawn sketching as an approach to observe how humans process an image from the perspective of shapes, or contours. We utilize a simple mapping procedure to represent the sketch data in image tracings as voxel grids. By leveraging Euclidean distance as a heuristic for measuring dissimilarity between sketches, we conduct a statistical analysis which suggests that shape attention re-prioritization occurs when participants are under time constraints. The effects of shape attention re-prioritization become more apparent as time constraints become more restrictive and sketches become more varied as a result. Future work includes developing other methods for sketch comparison, such as clustering individual strokes. It is yet to be determined if the full variation of shapes that people utilize for object recognition can be captured under a finite set of contour properties.


\bibliographystyle{apacite}

\setlength{\bibleftmargin}{.125in}
\setlength{\bibindent}{-\bibleftmargin}

\bibliography{CogSci_Template}

\end{document}